# Advancing Biomedicine with Graph Representation Learning: Recent Progress, Challenges, and Future Directions

Fang Li[1], Yi Nian[1], Zenan Sun[1], Cui Tao[1*]

[1] School of Biomedical Informatics, the University of Texas Health Science Center at Houston, Houston, TX, USA

**Corresponding author:**

Dr. Cui Tao, School of Biomedical Informatics, the University of Texas Health Science Center at Houston, 7000 Fannin, Suite 600, Houston, TX, 77030; Phone: 713-500-3981; Fax: 713-500-3929; e-mail: Cui.Tao@uth.tmc.edu

(Main text word count: 4,471, 31,518 characters with space, excluding Title page, Summary, Tables, Acknowledgements, and References; ≤ 34,000 characters including spaces)

## **Summary** (250, ≤ 250 words)

**Objectives:** Graph representation learning (GRL) has emerged as a pivotal field that has contributed significantly to breakthroughs in various fields, including biomedicine. The objective of this survey is to review the latest advancements in GRL methods and their applications in the biomedical field. We also highlight key challenges currently faced by GRL and outline potential directions for future research.

**Methods:** We conducted a comprehensive search of multiple databases, including PubMed, Web of Science, IEEE Xplore, and Google Scholar, to collect relevant publications from the past two years (2021-2022). The studies selected for review were based on their relevance to the topic and the publication quality.

**Results:** A total of 78 articles were included in our analysis. We identified three main categories of GRL methods and summarized their methodological foundations and notable models. In terms of GRL applications, we focused on two main topics: drug and disease. We analyzed the study frameworks and achievements of the prominent research. Based on the current state-of-the-art, we also discussed the challenges and future directions.

**Conclusions:** GRL methods applied in the biomedical field demonstrated several key characteristics, including the utilization of attention mechanisms to prioritize relevant features, a growing emphasis on model interpretability, and the combination of various techniques to improve model performance. However, there are challenges needed to be addressed, including mitigating model bias, accommodating the heterogeneity of large-scale knowledge graphs, and improving the availability of high-quality graph data. To fully leverage the potential of GRL, future efforts should prioritize these areas of research.

## 1. Introduction

The paradigm of evidence-based precision medicine has evolved toward profound utilization of large volumes of data, driven by the rapid development of high technologies and the exponential growth of biomedical data [1]. Graphs have emerged as a major form for describing and modeling ubiquitous real-life systems [2], owing to their ability to represent complex temporal and spatial relationships between entities [3]. A graph is a data structure that comprises nodes and edges, representinig entities and relationships between them, respectively.[4] Graph-structure data are pervasive in biomedicine and healthcare, representing information at molecular level (such as chemical structure [5] and gene regulatory network [6]),  patient level (such as comorbidity network [7]), and population level (such as epidemic network [8] and healthcare system [9]). Knowledge graphs (KGs), a type of heterogeneous graphs, are used to represent networked entities and relationships [10], where entities can denote real objects or theoretical concepts, and relationships indicate their associations. Moreover, entities and relationships are endowed with types and properties that accurately convey their semantics.[11] From a knowledge representation perspective, KGs can be regarded as relaxed-defined ontologies, which retain basic elements of ontology while disregarding restrictions of strictly-defined axioms—statements that are asserted to be true in the domain. Graphs, together with KGs, have many cutting-edge applications in healthcare, including drug repurposing [12], disease risk prediction [13], and protein-protein interaction (PPI) prediction [14], and can be used to generate new hypotheses that ultimately translate into clinically actionable outcomes.

Over the past two decades, machine learning (ML), specifically deep learning (DL), has been successful in vast healthcare scenarios, such as medical imaging and diagnostics [15], drug discovery [16], health insurance and fraud detection [17]. However, these techniques were mainly designed to process the Euclidean data, such as sequential electronic health records (EHRs), text, and image, and may not handle the non-Euclidean graph data directly [4] The distinct among the Euclidean and non-Euclidean data is the underlying geometry used to represent the data: the Euclidean geometry deals with flat, two-dimensional spaces, while the non-Euclidean geometry studies curved surfaces [18]. The key challenge in utilizing graph data in ML models is finding a way to represent graph structure that is easy for the models to learn.[19] Graph representation learning (GRL), which embeds raw graph data into a low-dimensional space while preserving graph topology and node properties [2], can make graph data more amenable to ML. As the frontier of graph algorithms, GRL has attracted significant interest from diverse fields, including computer science and biomedicine. Despite the remarkable success of GRL, it still faces many challenges, ranging from the theoretical understanding of methods to scalability and interpretability in real systems, and from the soundness of methodology to empirical performance in applications [20].

In this survey, we review the recent progress of GRL in biomedicine, focusing on both its methods and applications. We also identify the key challenges that GRL faces and discuss the potential opportunities for future research. Our goal is to provide insight into the direction of future efforts in this active and fast-growing research area.

## 2. Materials and Methods

This survey aimed to review the latest development in GRL research in the healthcare field during the past two years (2021–2022). To ensure that the analysis was in-depth and up-to-date, we conducted a thorough search for relevant articles in multiple databases, including PubMed, Web of Science, IEEE Xplore, Google Scholar, and arXiv. The study selection criteria focused on topic relevance and publication quality, with preference given to high-impact factor journals, top-tier conferences, and articles with large number of citations. By employing this rigorous approach, we included 78 most representative articles, encompassing both original research and reviews. The selected studies were of high quality, ensuring a robust understanding of the latest advancements in GRL in the biomedical field.

## 3. Advances in GRL Methods

Over the last decade, GRL has emerged as a critical and pervasive research area, greatly improving the efficiency and flexibility of representation learning [19]. There are two settings for graph learning patterns: transductive learning and inductive learning (or reasoning) [20,21]. Transductive learning involves observing all data, both training and test (with unknown labels), during training. The model learns from the observed training data and predicts the labels of the test data. On the other hand, inductive learning is more like the traditional supervised learning, where the model encounters only the training data when developing and then the learned model is applied on the test data which it has never seen before. Transductive learning can generate node embeddings for existing nodes or suggest new relations (edges) in a fixed graph, while the inductive learning has the generalizability to the new graphs.

In this section, we will explore three fundamental categories of GRL methods, based on the classification defined in a few studies [2,22,23]. **Table 1** presents the principles, characteristics and applicable tasks of these three GRL categories. Additionally, **Table 2** lists some notable GRL models that have been developed in recent years.

## 3.1. Shallow Node Embedding

The purpose of the shallow node embedding is to project nodes onto a latent space, which is a multi-dimensional vector space learned by a model based on the input data. This latent space serves as a summary of the local graph structure, and the original relations in the graph are then represented by the topological relationships of the embedded representation. Node embeddings are characterized by an encoding and decoding process [22], where the encoder maps each node to the latent embedding space, serveing as an embedding lookup table, while the decoder reconstructs a graph statistic for a pair of embedded nodes. The optimization of encoder and decoder is intended to minimize the loss between the decoded statistic and some node-based similarity metrics. Notable shallow node embedding methods include DeepWalk [24], node2vec [25], struc2vec [26], and LINE [27].

Shallow node embedding methods are relatively simple to implement and interpret. However, their transductive nature [28] makes them less suitable for inductive reasoning, where the graph structure may change or not be pre-defined [29]. Moreover, the shallow embedding methods only consider the topological structure of graph as input and generate the embedding of nodes or edges, without taking into account any associated node and edge attributes.

### 3.2. Graph Neural Networks

Graph neural networks (GNNs) are neural networks designed to operate on graph data.[20] By learning compact representations of graph elements, their attributes, and supervised labels (if any)[23], GNNs surpass shallow node embeddings in their ability to perform inductive reasoning and capture higher-order and nonlinear patterns through multi-hop propagation within several layers of neural message passing [30]. In contrast, shallow node embedding methods only generate node representations that can be combined with ML models to perform downstream tasks, while GNNs fuse both graph topology and attributes to perform end-to-end graph tasks.[2] Additionally, GNNs can optimize supervised signals and graph structure simultaneously, whereas shallow embeddings require a two-stage approach to achieve the same.[23]

Convolutional neural network (CNN) are among the most popular DL models used in the computer vision applications [31], and have shown exceptional performance in tasks such as object detection [32] and image analysis [33,34]. Although CNN are traditionally used for structured Euclidean data, such as image pixels or text sequences, the concept of convolution to learn local connections can be adapted to non-Euclidean graphs using spectral and spatial approaches [4]. In the spectral approach, graph information is transformed to the spectral domain using the graph Fourier transform and the eigendecomposition of the graph Laplacian [35], and convolution is performed on the graph spectrum. Graph convolutional network (GCN) [36], dual graph convolutional network (DGCN) [37], and Cluster-GCN [38] are typical GNN variants that use this approach. In the spatial approach, convolution is performed directly on the topological graph. However, unlike the convolution operation on image pixels, graph convolution lacks the weight-sharing property, and the size of the node's neighbors is not always the same.

To address these challenges in the spatial domain, several models have been developed, such as diffusion-convolutional neural network (DCNN) [39], graph sample and aggregate (GraphSAGE) [28], and mixture model network (MoNet) [40].

In addition to the aforementioned models, several other state-of-the-art neural networks are applicable to graphs. For instance, graph attention network (GAT) [41] employs the self-attention strategy to assign different weights to the neighbors of each node, allowing it to learn node representation on graphs with varying node degrees and enabling inductive learning. Gated-based models, such as Tree LSTM [42], gated graph neural network (GGNN) [43], and graph LSTM [44], utilize the gate mechanism to facilitate long-term information propagation. The gate operator allows information to be updated or discarded, which can help reduce the noise during the information propagation process.

Despite their success in a range of graph-based learning tasks, such as node classification, link prediction, and graph clustering, GNNs are often criticized for their lack of interpretability. As black box models, it can be challenging to discern how these networks make predictions or extract meaningful insights from the learned representations.[4] Additionally, the computational cost of GNNs can be prohibitive, particularly when dealing with large-scale biomedical graphs comprising millions of nodes and edges.[30] This constraint can limit their applicability in real-world scenarios where computational efficiency is critical.

## 3.3. Generative graph models

In recent years, generative graph models have emerged as a promising field in GRL. Unlike shallow embeddings and GNNs, which focus on learning embedding of existing graphs, generative graph models leverage graph characteristics, such as graph structure, node and edge information, to generate new graphs that possess similar properties to the original graph.

Two popular generative graph models are variational autoencoders (VAEs) and generative adversarial networks (GANs). VAEs utilizes stochastic variational inference to train an encoder and decoder that can generate graphs from a learned distribution based on a latent representation [45]. Models such as variational graph auto-encoder (VGAE) [46], GraphVAE [47], and junction tree variational autoencoder (JEVAE) [48] are examples of VAE-based approaches. On the other hand, GANs consist of a generator that produces fake samples and a discriminator that distinguishes between real and fake data [49]. The goal is to increase the likelihood of identifying the true samples as real and the reconstructed samples as fake. GraphGAN [50] and MolGAN [51] are two examples of GAN-based generative graph models. These generative graph models have shown great potential in biomedicine and healthcare field such as social network analysis [52], drug discovery [51], and protein structure construction [53].

However, there are still challenges to overcome, such as scalability and interpretability, to make the generative graph models more applicable to real-world scenarios. Additionally, generative graph models can be challenging to replicate, primarily due to their high sensitivity to the initial random seed used during the graph generation process.[54] As a result, even minor variations in the seed value can lead to significant differences in the generated graph structure, making it difficult to reproduce the same results.

# 4. Advances in GRL Applications for Biomedicine

From a graph machine learning perspective, research on GRL application can be divided into various tasks, including of node, triple, and graph classification, link (relation) prediction, node and graph clustering, and graph generation [22]. Considering the extensive range of GRL application studies available, we selected two crucial healthcare topics, namely drug and disease, to summarize some notaworthy studies. **Table 3** outlines the key components of GRL applications in research related to drug and disease.

## 4.1. Drug Development and Related Association Predictions

### 4.1.1 In Silico Drug repurposing

In the field of drug discovery and development, in silico drug repurposing, which involves the computational identification of new indications and targets for already marketed drugs [55], continues to be an attractive proposition. Drug repurposing relies on de-risked drugs, which have to potential to offer lower development costs and shorter development life-cycles [12]. The primary objective of drug repurposing is to identify candidate drugs that have a high probability of being associated with the therapeutic indication of interest [56]. This task can be framed as a link prediction challenge that aims to identify potential drug-target interaction (DTI) or drug-disease association with a high level of confidence.

The common approach for in silico drug repurposing involves predicting DTI. A drug target is a protein or other biomolecule (such as DNA, RNA, and peptide) that the drug binds to directly and is responsible for the drug's therapeutic efficacy [57]. Peng et al. [58] developed an end-to-end learning-based framework (EEG-DTI) that employed heterogeneous GCNs for DTI prediction.

Specifically, a heterogeneous network—comprising four types of nodes (drug, protein, disease, and side effect) and eight types of edges (interaction, association, and similarity)—was created by merging multiple biological networks. A three-layer GCN was then implemented to produce low-dimensional embeddings for drugs and proteins using information from their neighbors in the heterogeneous network. The drug and protein embeddings were concatenated, and the inner product was used to calculate the drug-protein interaction score (i.e., DTI prediction). Li *et al.* [59] introduced a multi-channel GCN and GAT-based framework (DTI-MGNN) for DTI prediction, utilizing topology graph (contextual representation), feature graph (semantic representation), and common representation of drug and protein pairs (DPPs). Xuan et al. [60] proposed a graph convolutional and variational autoencoder-based approach (GVDTI), which encoded multiple pairwise (drug-protein) representations, including topological representation, attribute representation, and attribute distribution. The three pairwise representations were then fused by convolutional and fully connected neural networks for DTI prediction. Similarly, Hsieh *et al.* [61] utilized variational graph autoencoders with GraphSAGE message passing to generate drug embeddings and selected the most potent drugs for COVID-19. Ding *et al.* [62] employed a relational graph convolutional network (RGCN) to predict the drug-protein interactions and further predict the blood-brain barrier permeability of drug molecules.

In addition to predicting DTIs, another approach to in silico drug repurposing is predicting drug-disease associations. A deep understanding of the mechanism of drug action (MDA) is required for drug repurposing, which is often explained through biological pathways—a series of biochemical and molecular steps to achieve a specific function or to produce a certain product. To capture MDA and identify the critical paths from drugs to diseases in the human body, Yang *et al.* [63] proposed an interpretable DL-based path-reasoning framework (iDPath) that employed a multilayer biological network and various modules, including a GCN module, an LSTM module, and two attention modules (the node and path attention). Experiments showed that iDPath could identify explicit critical paths that were consistent with clinical evidence. Nian *et al.* [64] utilized semantic triples in SemMedDB for KG construction and drug-disease link prediction. They filtered the most relevant semantic triples to Alzheimer's disease (AD) by a BERT-based classifier and some rule-based methods, and train graph embedding algorithms, such as TransE [65], DistMult [66], and ComplEx [67], to predict drug/chemical/food supplement candidates that may be helpful for AD treatment or prevention. Cai *et al.* [12] proposed a heterogeneous information fusion GCN approach (DRHGCN) for drug repurposing, which applied graph convolution operations to three networks (drug-drug similarity, disease-disease similarity, and drug-disease association networks) to learn the embedding of drugs and diseases. DRHGCN also designed inter- and intra-domain feature extraction modules, and a layer attention mechanism to further improve the prediction performance. Experiment results demonstrated that DRHGCN identified several novel approved drugs for AD and Parkinson's disease.

#### 4.1.2 Drug-drug interaction prediction

Drug-drug interactions (DDIs) occur when two or more drugs interact with each other and can alter the absorption of one or both drugs, leading to delayed, decreased, or enhanced effect. These interactions can have significant consequences, including synergistic effects, where the total effect of the drugs is greater than the sume of their individual effects, or antagonistic effects, where the drugs have opposing effects on the body, potentially reducing or blocking the effectiveness of one or more of the drugs [68]. Adverse effects can also occur as a result of DDIs.

Synergistic DDIs can be beneficial, particularly for cancer therapy, because they allow for the use of lower dose of chemothrapy drugs while maitaining or even enhancing their effectiveness. By contrast, antogonistic DDIs may reduce the efficacy of medications and require additional or alternative treatments.

Dai *et al.* [69] proposed a novel framework for DDI classification using an adversarial autoencoder-based embedding approach (AAEs). To address the challenge of generating high-quality negative samples, the authors utilized an autoencoder which learned to produce plausible negative triplets for the discriminator while minimizing reconstruction errors via the decoder component. The discriminator was trained on both the generated negative triplets and the original positive triplets to produce a robust and effective graph representation model. To tacke vanishing gradient issues in the discrete representation, the authors employed the Gumbel-Softmax relaxation and the Wasserstein distance for training the embedding model, which provided a more stable and efficient training process, allowing for improved performance and faster convergence.

Identifying synergistic anticancer drug combinations is a common scenario in synergistic DDI prediction. To address this, Wang *et al.* [70] proposed a DL-based framework called DeepDDS. The framework utlized a multilayer feedforward neural network (MLP) to obtain the feature embedding of gene expression profiles of the cancer cell line, and either GAT or GCN to obtain the feature embeddding of the drug (represented as a graph of molecular structure, from SMILE). The embedding vectors of the drug and the cell line were concatenated and fed into a multilayer fully connected network to predict the synergistic effect. The study also explored the interpretability of the GAT and found that the correlation matrix of atomic features revealed important chemical substructures of drugs. Yang *et al.* [71] developed GraphSynergy, a GCN-based framework for predicting synergistic DDI. GraphSynergy encoded the high-order topological relationships in the PPI network between proteins that were targeted by a pair of drugs and were associated with a specific cancer cell line. The pharmacological effects of drug combinations were evaluated by their therapy and toxicity scores. An attention component incorporated to capture the pivotal proteins that played a part in both the PPI network and biomolecular interactions between drug combinations and cancer cell lines.

Bang *et al.* [72] developed a graph feature attention network (GFAN) for predicting polypharmacy side effects with enhanced interpretability. Polypharmacy refers to the concurrent use of two or more different drugs. The GFAN model emphasized target genes differently for each side-effect prediction, making it capable of sensitively extracting target genes and providing interpretability. The experiments conducted by the authors showed that the GFAN model was effective in predicting polypharmacy side effects.

## 4.2. Disease and Related Association Predictions

### 4.2.1 Disease prediction

Disease prediction using EHRs has become an area of significant research interest due to their increasing availability. EHR-based prediction and classification include predicting clinical risks, disease subtyping, and chronic disease onset, among others. However, conventional ML approaches rely heavily on abundant data to train the models, which can impede their performance in predicting rare diseases with severe data scarcity. Additionally, most existing disease prediction approaches rely on sequential EHRs, making it difficult to handle new patients without historical records.

To overcome these challenges, Sun *et al.* [13] proposed a GNN-based graph encoder that leveraged GATs and graph isomorphism networks (GINs) to learn highly representative node embedding for patients. This approach utilized both the external knowledge base (human phenotype ontology) and patients' EHRs represented in the graph structure. The well-learned graph encoder can inductively infer the embeddings for a new patient, enabling prediction of both general and rare diseases. The study demonstrated promising results in addressing the scarcity of training data for rare disease prediction.

EHRs contain tens of thousands of medical concepts that are implicitly connected. A feasible approach to improving EHR representation learning is to associate relevant medical concepts and leverage these connections. To this end, Zhu *et al.* proposed a variationally regularized encoder-decoder graph neural network (VGNN) for EHRs that achieved robustness in graph structure learning by regularizing node representations [73]. Another approach to leveraging connections among medical concepts is to exploit diagnoses as relational information by connecting similar patients in a graph. Rocheteau *et al.* [74] proposed such a strategy by designing an LSTM-GNN model for patient outcome prediction. The model extracted temporal features using LSTM and extracted the patient neighborhood information using GNNs. The results showed that the LSTM-GNN outperformed the LSTM-only baseline on length of stay prediction tasks on the eICU database, indicating that exploiting information from neighboring patient cases using GNNs is a promising research direction in EHRs-based supervised learning.

Xia *et al.* [75] developed a medical conversational question-answering system that utilized a multi-modal clinical KG as its knowledge base to support entity reasoning, such as diseases, medical examinations, and drugs based on the patient's symptoms collected by the system. The system is equipped with advanced natural language processing (NLP) techniques such as contrastive learning, prompt, bi-directional encoder, and autoregressive decoder, which helped to achieve state-of-the-art performance. With the multi-modal clinical KG and advanced NLP techniques, the system can answer complex medical questions in a conversational manner, making it a promising tool for assisting clinical decision-making and patient care.

#### 4.2.2 Disease-protein/RNA association prediction

MicroRNAs (miRNAs) are crucial in the development of human complex diseases. Discovering the associations between miRNAs and diseases is essential for both basic and translational medicine. To address this, Tang *et al.* [76] developed a multi-view multichannel attention GCN (MMGCN) to predict potential miRNA-disease associations. This approach utilized a GCN encoder that took multiple similarity graphs of miRNA and disease as input, fused their neighbor information, and generated their embeddings under different views (i.e., graphs). The multichannel attention mechanism on miRNA and disease prioritized important channel embedding and produce normalized channel attention features. Additionally, a CNN combiner was used to convolve the multichannel attention features of miRNA and disease, respectively, to generate corresponding representations for association prediction. The MMGCN approach is a promising method for predicting miRNA-disease associations, which could aid in the development of novel therapies for complex human diseases.

#### 4.2.3 Disease-microbe association prediction

Human microbes play a critical role in a wide range of complex diseases and have become a new target in precision medicine. In silico identification of microbe-disease associations can provide insights into understanding the pathogenic mechanism of complex human disease and facilitate screening candidate targets for drug development. To this end, Long *et al.* [77] proposed a GAT-based framework, GATMDA, for human microbe-disease association prediction. The framework leveraged multiple similarity-based graphs to construct input features for microbes and diseases. GAT with talking-heads was employed to learn the representations of microbe and disease nodes. To filter out noises and focus on more important neighbors, a bi-interaction aggregator was utilized to enforce representation aggregation of similar neighbors. Finally, the inductive matrix completion (IMC) was combined to reconstruct a bipartite graph to predict microbe-disease associations. The proposed framework showed promising results in identifying potential microbe-disease associations, highlighting the potential for using GRL to facilitate precision medicine.

## 5. Challenges and Future Directions

### 5.1. Bias

Bias refers to the presence of prejudice or favoritism toward an individual or a group based on their inherent or acquired characteristics when making a decision [78]. When an algorithm's decisions favor or disfavor a specific group disproportionately, it is said to be biased. In the biomedical domain, graph algorithms are used in many applications, but they can be susceptible to bias. For example, in the case of melanoma detection, ML models are trained using images from fair-skinned populations, primarily from the United States, Europe, and Australia. Consequently, these models may perform poorly in detecting lesions from individuals with different skin colors, indicating inherent bias in the model [79].

The issue of bias in algorithms can be mitigated at three stages of the ML pipeline [80]. At the pre-processing stage, bias can be addressed by generating non-discriminatory labeled data and obtaining fair data representations. However, generating non-discriminatory data can be difficult, especially for health records that ofen include sensitive features such as sexual identity, race, and other social determinants of health. At the in-processing stage, algorithms could be modified to avoid bias, such as by changing the sampling strategy and adding regularization terms to the training process. For example, FairWalk [81], a graph embedding algorithm derived from node2vec, partitions neighbors into groups based on their sensitive attribute values and gives each group an equal probability of being chosen, thereby removing biases such as gender and race to a large extent. However, addressing multiple cross-attribute biases in networks with richer subgroup fairness still poses a challenge. At the post-processing stage, bias can be addressed by altering the classification threshold to ensure model fairness [82]. To the best of our knowledge, there are no GNN-based studies yet that alleviate the bias problem at the post-processing stage in the biomedical field. In the future, it would be advantegeous to develop a sensitive information-oriented graph ML framework that integrates different modules for a variety of subgroups and considers background information such as the methods of data acquisition, training data creation, and model training to address bias in biomedical data.

## 5.2. Interpretability

Graph algorithms are widely used for their accuracy in predicting outcomes, but they often lack of interpretability [83], making them difficult to trust or use safely in sensitive domains such as healthcare. In order to ensure the transparency and trustworthiness of graph algorithms, it is important to provide both accurate predictions and human-intelligible explanations. To this end, several types of GNN explanation methods have been proposed to explain node and graph classification tasks. Here we summarized three categories: mask-based methods, such as GNNExplainer [84], PGExplainer [85], and ZORRO [86]; perturbation-based methods, such as PGM-Explainer [87]; and generative model-based methods, such as XGNN [88]. Mask-based approaches generate a new graph by combining masks with the original features/edges/nodes, enabling them to capture important feature/edge/node information during backpropagation. Perturbation-based methods filter out unimportant features using data sampling, then fit an explainable small model like a probabilistic graphical model (PGM) on filtered data for topological explanation. Generative model-based methods generate small explainable subgraphs in a node-by-node way. For instance, XGNN [88] uses a reinforcement learning framework to learn the probability of growing from a node to a subgraph for the explanation.

One major challenge of using explainability methods is determining how to assess their effectiveness. To address this issue, many studies have used synthetic data and real-world datasets, such as MUTAG [89] and MNIST [90], to validate their models. However, these validation datasets are often relatively small and straightforward, raising concerns about whether these explainable graph algorithms can be generalized to large-scale biomedical graph data. Recently, Agarwal et al. [91] proposed an approach for evaluating explainability of GNNs. They developed a synthetic graph data generator, SHAPEGGEN, that can generate a variety of benchmark datasets and provide ground-truth explanations.

Furthermore, future efforts could focus on exploring novel training strategies to explaining other tasks, such as link prediction, beyond current approaches for addressing interpretability issues for classification tasks (i.e., node and graph classification) [92].

Additionally, incorporating edge-based explanations, in addition to node-based explanations, would be beneficial in assisting human experts [93].

## 5.3. Heterogeneity

Biomedical graph data, particularly KG, often contain different classes of nodes, such as diseases, drugs, and genes. Graph ML tasks have shown that GNNs perform better than traditional methods on diverse graph data. However, recent studies suggested that GNNs, such as GCNs, may have inferior performance in heterogeneous graphs than in homogenous ones.[94] To address this issue, methods such as heterogeneous graph transformer (HGT) [95] and heterogeneous graph attention network (HAN) [96] have been developed. HGT utilizes the meta-relations to parameterize weight matrices for heterogeneous mutual attention, message passing, and propagations [95]. HAN, on the other hand, leverages both node-level and semantic-level attention to simultaneously consider the importance of nodes and meta-paths [96]. However, these methods have only been validated on the datasets with less than five types of nodes, whereas common biomedical KGs tend to be more heterogeneous and complex, with greater scale. Therefore, newer graph ML models that are specifically designed for large-scale heterogeneous biomedical KGs are still necessary.

## 5.4 Availability of high-quality graph data

GRL heavily relies on the availability of high-quality graph data or KGs. The performance of GRL algorithms is directly linked to the quality and quantity of the input data. Incomplete or low-quality graph data can result in inaccurate embeddings, which can adversely affect the performance of downstream tasks. Moreover, GRL algorithms are computationally intensive and require large mounts of data to train effectively. Therefore, if the available graph data are limited, the performance of GRL algorithms can be severely impacted. Additionally, since KGs are often constructed manually or semi-automatically by domain experts, the process can be time-consuming, costly, and prone to errors. This can limit the amount of high-quality graph data available, further restricting the applicability of GRL.

The availability of high-quality graph data or KGs is essential for GRL algorithms to be effective and scalable. Therefore, improving the quality and quantity of available graph data is crucial to unlock the full potential of GRL in biomedical applications.

## 6. Conclusions

In this survey, we have highlighted the significant advancements made in the field of GRL in biomedicine. GRLs techniques have been extensively utilized to bridge major gaps in healthcare, enabling researchers to unravel complex disease mechanisms, accelarate drug discovery, and enhance personalized disease prediction and management. These breakthroughs are also a result of interdisciplinary collaborations among computer scientists, biologists, and medical professionals, and their concerted efforts to integrate knowledge from diverse fields. Looking ahead, we anticipate that the development of more robust, interpretable, and trustworthy GRL algorithms, along with the availability of high-quality graph data, particulary well-curated KGs, will continue to play a critical role in advancing precision medicine. As GRL techniques continue to mature, they hold immense promise for boosting precision medicine by harnessing vast amounts of graph data in a meaningful and interpretable way.

## Acknowledgments

The authors were supported by grants from the National Institutes of Health under Award Numbers RF1AG072799, R01AI130460, R56AG074604, and the American Heart Association under Award No.19GPSGC35180031.

**Table 1 Principles, characteristics and applicable tasks of three categories of GRL methods**

| Category | Principles | Characteristics | | Representative methods | Biomedical Applications |
|---|---|---|---|---|---|
| | | Strengths | Limitations | | |
| Shallow node embedding | Project nodes into a latent space that summarizes their graph positions and local graph neighborhoods. | ● Easy to implement and interpret | ● Does not include node features<br>● Rely on embedding lookup tables and are transductive reasoning | DeepWalk [24], node2vec [25], struc2vec [26], LINE [27] | Node classification (e.g., protein function prediction), link prediction (e.g., drug repurposing) |
| Graph neural networks | Leverage neural networks to learn compact representations of graph topology and node attributes. | ● Can be used for inductive reasoning<br>● Capture higher-order and nonlinear patterns through multi-hop propagation | ● Computationally expensive on large graphs<br>● Lack of interpretability | GCN [36], DGCN [37], Cluster GCN [38], DCNN [39], GraphSage [28], MoNet [40], GAT [41], GGNN [43], Graph LSTM [44] | Node classification, linke prediction, graph prediction |
| Generative graph models | Employ graph neural networks to learn representations and generate realistic graph structures. | ● Generate graphs with different characteristics and properties | ● Can be difficult to reproduce the same result due to the high sensitivity to the initial random seeds for graph generation | VGAE [46], GraphVAE [47], JEVAE [48], MolGAN [51], GraphGAN [50] | Molecular graph generation |

**Table 2 Recent Notable GRL Models**

| Model Name | Core Components | Characteristics and Strengths | Reported Tasks |
|---|---|---|---|
| SIGN (Scalable inception graph neural network) [97] | GCN | • Used different types and sizes of graph convolutional filters to substitute graph sampling;<br>• Resulted in fast training and inference on complex graph | Node classification |
| AGE (Adaptive graph encoder) [98] | GCN, Laplacian smoothing, adaptive learning | • Designed a Laplacian smoothing filter for GCN to get smoothed features;<br>• Applied an adaptive encoder to avoid noise in node embedding | Node clustering, link prediction |
| GraphSAINT (Graph sampling-based inductive learning method) [99] | GCN, GAT | • Sampled small and complete subgraphs across GCN layers to overcome neighbor explosion problem;<br>• Improved accuracy while lowering training time on large graphs | Node classification, link prediction |
| scGNN (Single-cell graph neural network) [100] | GCN, heterogeneous graph, left-truncated mixed Gaussian modeling | ● Employed multi-modal autoencoders to iteratively learn graph representation until converging;<br>● Used a left-truncated mixed Gaussian model to regularize the learned node embedding;<br>● Provided a valuable framework for general scRNA-Seq analyses | Node classification, node clustering, link prediction |
| MAGNN (Metapath aggregated graph neural network) [101] | GAT, heterogeneous graph | ● Applied attention mechanism to aggregate information of intermediate nodes from each metapath;<br>● Maintained both structural and semantic features for heterogeneous graph embedding | Node classification, node clustering, link prediction |
| HGT (Heterogeneous graph transformer) [102] | GAT, heterogeneous graph, dynamic graph | ● Calculated attention over meta-relations to model heterogeneous graphs;<br>● Integrated all the relations from different timestamps and measured their structural temporal dependencies to handle dynamicity;<br>● Improved various downstream tasks on the Web-scale Open Academic Graph | Node classification |
| GCA (Graph contrastive learning with adaptive augmentation) [103] | Contrastive representation learning | ● Applied an adaptive argumentation to generate the topological and semantic view of input graph with graph contrastive learning method;<br>● Adaptively removed unimportant relations and added noise to irrelevant features to learn implicit semantic information as well as significant graph structures | Node classification |
| GraphAF (flow-based autoregressive model for graph generation) [104] | GCN, autoregressive model | ● Used an autoregressive model to dynamically generate nodes and edges, and a relational GCN to learn node representations;<br>● Allowed parallel computation to get exact data likelihood, which further improved training efficiency | Graph generation |
| STGD-VAE (SpatioTemporal graph disentangled variational auto-encoder) [105] | VAE, dynamic graphs, Bayesian model | ● Considered time-invariant geometric factors, graph factors, and spatial-graph joint factors in the disentangled representation learning to generate new spatiotemporal graphs;<br>● The first general deep generative model framework for dynamic (spatiotemporal) graphs | Graph generation |

**Table 3 Core Elements of KG Applications in Drug and Disease-Related Research**

| Domain | Research question | Predicted node pairs | Possible input |
|---|---|---|---|
| Drug | Drug repurposing [58–61], drug development [63] | Drug-disease, Drug-target/gene/protein | Molecular graph of chemical;<br>drug-drug similarity matrix;<br>drug-drug interaction network； drug-gene/protein/disease association network |
| | Drug-drug interaction [69–71] | Drug-drug, drug-drug-cell line | |
| | Drug side effect [72] | Drug-disease | |
| Disease | Disease prediction [13,73,74] | Disease | Clinical feature network |
| | Genetic association [76] | Disease - DNA/miRNA/lncRNA | Disease-disease similarity matrix; gene-gene similarity matrix;<br>Gene expression profile;<br>disease-gene/protein association<br>Protein-protein interaction network |
| | Pathogenic association [77] | Disease - microbe/pathogen | Microbe-microbe similarity matrix;<br>Disease-microbe association |

## References

1 *Santos A, Colaço AR, Nielsen AB, Niu L, Strauss M, Geyer PE, Coscia F, Albrechtsen NJW, Mundt F, Jensen LJ, Mann M.* A knowledge graph to interpret clinical proteomics data. Nat Biotechnol 2022; 40: 692–702

2 *Yi H-C, You Z-H, Huang D-S, Kwoh CK.* Graph representation learning in bioinformatics: trends, methods and applications. Brief Bioinform 2022; 23 Available from: http://dx.doi.org/10.1093/bib/bbab340

3 *Leser U, Trißl S.* Graph Management in the Life Sciences. In: Liu L, Özsu MT (eds.). Encyclopedia of Database Systems. Boston, MA: Springer US, 2009: 1266–1271

4 *Zhou J, Cui G, Hu S, Zhang Z, Yang C, Liu Z, Wang L, Li C, Sun M.* Graph neural networks: A review of methods and applications. AI Open 2020; 1: 57–81

5 *David L, Thakkar A, Mercado R, Engkvist O.* Molecular representations in AI-driven drug discovery: a review and practical guide. J Cheminform 2020; 12: 56

6 *Weighill D, Guebila MB, Lopes-Ramos C, Glass K, Quackenbush J, Platig J, Burkholz R.* Gene regulatory network inference as relaxed graph matching. Proc Conf AAAI Artif Intell 2021; 35: 10263–10272

7 *Khan A, Uddin S, Srinivasan U.* Comorbidity network for chronic disease: A novel approach to understand type 2 diabetes progression. Int J Med Inform 2018; 115: 1–9

8 *Gómez A, Oliveira G.* New approaches to epidemic modeling on networks. Sci Rep 2023; 13: 468

9 *Britto MT, Fuller SC, Kaplan HC, Kotagal U, Lannon C, Margolis PA, Muething SE, Schoettker PJ, Seid M.* Using a network organisational architecture to support the development of Learning Healthcare Systems. BMJ Qual Saf 2018; 27: 937–946

10 *Chandak P, Huang K, Zitnik M.* Building a knowledge graph to enable precision medicine. Sci Data 2023; 10: 67

11 *Ji S, Pan S, Cambria E, Marttinen P, Yu PS.* A Survey on Knowledge Graphs: Representation, Acquisition, and Applications. IEEE Transactions on Neural Networks and Learning Systems 2022; 33: 494–514 Available from: http://dx.doi.org/10.1109/tnnls.2021.3070843

12 *Cai L, Lu C, Xu J, Meng Y, Wang P, Fu X, Zeng X, Su Y.* Drug repositioning based on the heterogeneous information fusion graph convolutional network. Brief Bioinform 2021; 22 Available from: http://dx.doi.org/10.1093/bib/bbab319

13 *Sun Z, Yin H, Chen H, Chen T, Cui L, Yang F.* Disease Prediction via Graph Neural Networks. IEEE J Biomed Health Inform 2021; 25: 818–826

14 *Yuan Q, Chen J, Zhao H, Zhou Y, Yang Y.* Structure-aware protein–protein interaction site prediction using deep graph convolutional network. Bioinformatics 2021; 38: 125–132 15 *Aggarwal R, Sounderajah V, Martin G, Ting DSW, Karthikesalingam A, King D, Ashrafian H, Darzi A.* Diagnostic accuracy of deep learning in medical imaging: a systematic review and meta-analysis. NPJ Digit Med 2021; 4: 65

16 *Askr H, Elgeldawi E, Aboul Ella H, Elshaier YAMM, Gomaa MM, Hassanien AE.* Deep learning in drug discovery: an integrative review and future challenges. Artif Intell Rev 2022; 1–63

17 *Zhang G, Zhang X, Bilal M, Dou W, Xu X, Rodrigues JJPC.* Identifying fraud in medical insurance based on blockchain and deep learning. Future Gener Comput Syst 2022; 130: 140–154

18 *Singh PK.* Data with Non-Euclidean Geometry and its Characterization. Journal of Artificial Intelligence and Technology 2021; Available from: https://ojs.istp-press.com/jait/article/view/59

19 *Hamilton WL, Ying R, Leskovec J.* Representation Learning on Graphs: Methods and Applications. arXiv [csSI] 2017; Available from: http://arxiv.org/abs/1709.05584

20 Graph Neural Networks: Foundations, Frontiers, and Applications. Springer Nature Singapore

21 *Yang Z, Cohen WW, Salakhutdinov R.* Revisiting semi-supervised learning with graph embeddings. In: Proceedings of the 33rd International Conference on International Conference on Machine Learning - Volume 48. JMLR.org, 2016: 40–48

22 *Hamilton WL.* Graph Representation Learning. Synthesis Lectures on Artificial Intelligence and Machine Learning 2020; 14: 1–159

23 *Li MM, Huang K, Zitnik M.* Graph representation learning in biomedicine and healthcare. Nature Biomedical Engineering 2022; Available from: http://dx.doi.org/10.1038/s41551-022-00942-x

24 *Perozzi B, Al-Rfou R, Skiena S.* DeepWalk: Online Learning of Social Representations. arXiv [csSI] 2014; Available from: http://arxiv.org/abs/1403.6652

25 *Grover A, Leskovec J.* node2vec: Scalable Feature Learning for Networks. KDD 2016; 2016: 855–864

26 *Ribeiro LFR, Savarese PHP, Figueiredo DR.* struc2vec: Learning Node Representations from Structural Identity. arXiv [csSI] 2017; Available from: http://arxiv.org/abs/1704.03165

27 *Tang J, Qu M, Wang M, Zhang M, Yan J, Mei Q.* LINE: Large-scale Information Network Embedding. arXiv [csLG] 2015; Available from: http://arxiv.org/abs/1503.03578

28 *Hamilton WL, Ying R, Leskovec J.* Inductive representation learning on large graphs. arXiv [csSI] 2017; Available from: https://proceedings.neurips.cc/paper/6703-inductive-representation-learning-on-large-graphs

29 *Chami I, Abu-El-Haija S, Perozzi B, Ré C, Murphy K.* Machine learning on graphs: A model and comprehensive taxonomy. arXiv [csLG] 2020; Available from: https://www.jmlr.org/papers/volume23/20-852/20-852.pdf

30 *Morris C, Ritzert M, Fey M, Hamilton WL, Lenssen JE, Rattan G, Grohe M.* Weisfeiler and Leman Go Neural: Higher-Order Graph Neural Networks. AAAI 2019; 33: 4602–4609

31 *Alzubaidi L, Zhang J, Humaidi AJ, Al-Dujaili A, Duan Y, Al-Shamma O, Santamaría J, Fadhel MA, Al-*

*Amidie M, Farhan L.* Review of deep learning: concepts, CNN architectures, challenges, applications, future directions. J Big Data 2021; 8: 53

32 *Zhao Z-Q, Zheng P, Xu S-T, Wu X.* Object Detection With Deep Learning: A Review. IEEE Trans Neural Netw Learn Syst 2019; 30: 3212–3232

33 *Yao G, Lei T, Zhong J.* A review of Convolutional-Neural-Network-based action recognition. Pattern Recognit Lett 2019; 118: 14–22

34 *Ker J, Wang L, Rao J, Lim T.* Deep Learning Applications in Medical Image Analysis. IEEE Access 2018; 6: 9375–9389

35 *Shuman DI, Narang SK, Frossard P, Ortega A, Vandergheynst P.* The Emerging Field of Signal Processing on Graphs: Extending High-Dimensional Data Analysis to Networks and Other Irregular Domains. arXiv [csDM] 2012; Available from: http://arxiv.org/abs/1211.0053

36 *Kipf TN, Welling M.* Semi-Supervised Classification with Graph Convolutional Networks. arXiv [csLG] 2016; Available from: http://arxiv.org/abs/1609.02907

37 *Zheng Y, Gao C, Chen L, Jin D, Li Y.* DGCN: Diversified Recommendation with Graph Convolutional Networks. arXiv [csIR] 2021; Available from: http://arxiv.org/abs/2108.06952

38 *Chiang W-L, Liu X, Si S, Li Y, Bengio S, Hsieh C-J.* Cluster-GCN: An Efficient Algorithm for Training Deep and Large Graph Convolutional Networks. arXiv [csLG] 2019; Available from: http://arxiv.org/abs/1905.07953

39 *Li Y, Yu R, Shahabi C, Liu Y.* Diffusion Convolutional Recurrent Neural Network: Data-Driven Traffic Forecasting. arXiv [csLG] 2017; Available from: http://arxiv.org/abs/1707.01926

40 *Monti F, Boscaini D, Masci J, Rodolà E, Svoboda J, Bronstein MM.* Geometric deep learning on graphs and manifolds using mixture model CNNs. arXiv [csCV] 2016; Available from: http://arxiv.org/abs/1611.08402

41 *Veličković P, Cucurull G, Casanova A, Romero A, Liò P, Bengio Y.* Graph Attention Networks. arXiv [statML] 2017; Available from: http://arxiv.org/abs/1710.10903

42 *Tai KS, Socher R, Manning CD.* Improved Semantic Representations From Tree-Structured Long Short-Term Memory Networks. arXiv [csCL] 2015; Available from: http://arxiv.org/abs/1503.00075

43 *Li Y, Tarlow D, Brockschmidt M, Zemel R.* Gated Graph Sequence Neural Networks. arXiv [csLG] 2015; Available from: http://arxiv.org/abs/1511.05493

44 *Peng N, Poon H, Quirk C, Toutanova K, Yih W-T.* Cross-Sentence N-ary Relation Extraction with Graph LSTMs. arXiv [csCL] 2017; Available from: http://arxiv.org/abs/1708.03743

45 *Kingma DP, Welling M.* Auto-Encoding Variational Bayes. arXiv [statML] 2013; Available from: http://arxiv.org/abs/1312.6114v11

46 *Kipf TN, Welling M.* Variational Graph Auto-Encoders. arXiv [statML] 2016; Available from: http://arxiv.org/abs/1611.07308

47 *Simonovsky M, Komodakis N.* GraphVAE: Towards Generation of Small Graphs Using Variational Autoencoders. arXiv [csLG] 2018; Available from: http://arxiv.org/abs/1802.03480

48 *Jin W, Barzilay R, Jaakkola T.* Junction Tree Variational Autoencoder for Molecular Graph Generation. arXiv [csLG] 2018; Available from: http://arxiv.org/abs/1802.04364

49 *Goodfellow IJ, Pouget-Abadie J, Mirza M, Xu B, Warde-Farley D, Ozair S, Courville A, Bengio Y.* Generative Adversarial Networks. arXiv [statML] 2014; Available from: http://arxiv.org/abs/1406.2661

50 *Wang H, Wang J, Wang J, Zhao M, Zhang W, Zhang F, Xie X, Guo M.* GraphGAN: Graph Representation Learning with Generative Adversarial Nets. arXiv [csLG] 2017; Available from: http://arxiv.org/abs/1711.08267

51 *De Cao N, Kipf T.* MolGAN: An implicit generative model for small molecular graphs. arXiv [statML] 2018; Available from: http://arxiv.org/abs/1805.11973

52 *Grover A, Zweig A, Ermon S.* Graphite: Iterative Generative Modeling of Graphs. arXiv [statML] 2018; Available from: http://arxiv.org/abs/1803.10459

53 *Anand N, Huang P.* Generative modeling for protein structures. Adv Neural Inf Process Syst 2018; 31 Available from: https://proceedings.neurips.cc/paper/7978-generative-modeling-for-protein-structures

54 *Guo X, Zhao L.* A Systematic Survey on Deep Generative Models for Graph Generation. IEEE Trans Pattern Anal Mach Intell 2022; PP Available from: http://dx.doi.org/10.1109/TPAMI.2022.3214832

55 *Park K.* A review of computational drug repurposing. Transl Clin Pharmacol 2019; 27: 59–63

56 *Su C, Hou Y, Wang F.* GNN-based Biomedical Knowledge Graph Mining in Drug Development. In: Wu L, Cui P, Pei J, Zhao L (eds.). Graph Neural Networks: Foundations, Frontiers, and Applications. Singapore: Springer Nature Singapore, 2022: 517–540

57 *Santos R, Ursu O, Gaulton A, Bento AP, Donadi RS, Bologa CG, Karlsson A, Al-Lazikani B, Hersey A, Oprea TI, Overington JP.* A comprehensive map of molecular drug targets. Nat Rev Drug Discov 2017; 16: 19–34

58 *Peng J, Wang Y, Guan J, Li J, Han R, Hao J, Wei Z, Shang X.* An end-to-end heterogeneous graph representation learning-based framework for drug–target interaction prediction. Brief Bioinform 2021; 22: bbaa430

59 *Li Y, Qiao G, Wang K, Wang G.* Drug–target interaction predication via multi-channel graph neural networks. Brief Bioinform 2021; 23: bbab346

60 *Xuan P, Fan M, Cui H, Zhang T, Nakaguchi T.* GVDTI: graph convolutional and variational autoencoders with attribute-level attention for drug–protein interaction prediction. Brief Bioinform 2021; 23: bbab453

61 *Hsieh K, Wang Y, Chen L, Zhao Z, Savitz S, Jiang X, Tang J, Kim Y.* Drug repurposing for COVID-19 using graph neural network and harmonizing multiple evidence. Sci Rep 2021; 11: 23179 62 *Ding Y, Jiang X, Kim Y.* Relational graph convolutional networks for predicting blood–brain barrier penetration of drug molecules. Bioinformatics 2022; 38: 2826–2831

63 *Yang J, Li Z, Wu WKK, Yu S, Xu Z, Chu Q, Zhang Q.* Deep learning identifies explainable reasoning paths of mechanism of action for drug repurposing from multilayer biological network. Brief Bioinform 2022; Available from: http://dx.doi.org/10.1093/bib/bbac469

64 *Nian Y, Hu X, Zhang R, Feng J, Du J, Li F, Bu L, Zhang Y, Chen Y, Tao C.* Mining on Alzheimer's diseases related knowledge graph to identity potential AD-related semantic triples for drug repurposing. BMC Bioinformatics 2022; 23: 407

65 *Bordes A, Usunier N, Garcia-Duran A, Weston J, Yakhnenko O.* Translating embeddings for modeling multi-relational data. Adv Neural Inf Process Syst 2013; 26 Available from: https://proceedings.neurips.cc/paper/5071-translating-embeddings-for-modeling-multi-rela

66 *Yang B, Yih W-T, He X, Gao J, Deng L.* Embedding Entities and Relations for Learning and Inference in Knowledge Bases. arXiv [csCL] 2014; Available from: http://arxiv.org/abs/1412.6575

67 *Trouillon T, Welbl J, Riedel S, Gaussier E, Bouchard G.* Complex Embeddings for Simple Link Prediction. In: Balcan MF, Weinberger KQ (eds.). Proceedings of The 33rd International Conference on Machine Learning. New York, New York, USA: PMLR, 20--22 Jun 2016: 2071–2080

68 Drug Synergism. Available from: https://clinicalinfo.hiv.gov/en/glossary/drug-synergism

69 *Dai Y, Guo C, Guo W, Eickhoff C.* Drug–drug interaction prediction with Wasserstein Adversarial Autoencoder-based knowledge graph embeddings. Brief Bioinform 2020; 22: bbaa256

70 *Wang J, Liu X, Shen S, Deng L, Liu H.* DeepDDS: deep graph neural network with attention mechanism to predict synergistic drug combinations. Brief Bioinform 2022; 23 Available from: http://dx.doi.org/10.1093/bib/bbab390

71 *Yang J, Xu Z, Wu WKK, Chu Q, Zhang Q.* GraphSynergy: a network-inspired deep learning model for anticancer drug combination prediction. J Am Med Inform Assoc 2021; 28: 2336–2345

72 *Bang S, Ho Jhee J, Shin H.* Polypharmacy Side effect Prediction with Enhanced Interpretability Based on Graph Feature Attention Network. Bioinformatics 2021; Available from: http://dx.doi.org/10.1093/bioinformatics/btab174

73 *Zhu W, Razavian N.* Variationally regularized graph-based representation learning for electronic health records. In: Proceedings of the Conference on Health, Inference, and Learning. New York, NY, USA: Association for Computing Machinery, 2021: 1–13

74 *Rocheteau E, Tong C, Veličković P, Lane N, Liò P.* Predicting Patient Outcomes with Graph Representation Learning. arXiv [csLG] 2021; Available from: http://arxiv.org/abs/2101.03940

75 *Xia F, Li B, Weng Y, He S, Liu K, Sun B, Li S, Zhao J.* LingYi: Medical Conversational Question Answering System based on Multi-modal Knowledge Graphs. arXiv [csCL] 2022; Available from: http://arxiv.org/abs/2204.09220

76 *Tang X, Luo J, Shen C, Lai Z.* Multi-view Multichannel Attention Graph Convolutional Network for

miRNA–disease association prediction. Brief Bioinform 2021; 22: bbab174

77 *Long Y, Luo J, Zhang Y, Xia Y.* Predicting human microbe–disease associations via graph attention networks with inductive matrix completion. Brief Bioinform 2020; 22: bbaa146

78 *Mehrabi N, Morstatter F, Saxena N, Lerman K, Galstyan A.* A Survey on Bias and Fairness in Machine Learning. ACM Comput Surv 2021; 54: 1–35

79 *Adamson AS, Smith A.* Machine Learning and Health Care Disparities in Dermatology. JAMA Dermatol 2018; 154: 1247–1248

80 *Dai E, Wang S.* Say No to the Discrimination: Learning Fair Graph Neural Networks with Limited Sensitive Attribute Information. In: Proceedings of the 14th ACM International Conference on Web Search and Data Mining. New York, NY, USA: Association for Computing Machinery, 2021: 680–688

81 *Rahman T, Surma B, Backes M, Zhang Y.* Fairwalk: Towards Fair Graph Embedding. 2019 Available from: https://publications.cispa.saarland/2933/

82 *Lohia PK, Ramamurthy KN, Bhide M, Saha D, Varshney KR, Puri R.* Bias Mitigation Post-processing for Individual and Group Fairness. arXiv [csLG] 2018; Available from: http://arxiv.org/abs/1812.06135

83 *Masoomi A, Hill D, Xu Z, Hersh CP, Silverman EK, Castaldi PJ, Ioannidis S, Dy J.* Explanations of Black-Box Models based on Directional Feature Interactions. 2022; Available from: https://openreview.net/pdf?id=45Mr7LeKR9

84 *Ying R, Bourgeois D, You J, Zitnik M, Leskovec J.* GNNExplainer: Generating Explanations for Graph Neural Networks. Adv Neural Inf Process Syst 2019; 32: 9240–9251

85 *Luo D, Cheng W, Xu D, Yu W, Zong B, Chen H, Zhang X.* Parameterized Explainer for Graph Neural Network. arXiv [csLG] 2020; 19620–19631 Available from: https://proceedings.neurips.cc/paper/2020/hash/e37b08dd3015330dcbb5d6663667b8b8-Abstract.html

86 *Funke T, Khosla M, Anand A.* Hard Masking for Explaining Graph Neural Networks. 2021; Available from: https://openreview.net/pdf?id=uDN8pRAdsoC

87 *Vu MN, Thai MT.* PGM-explainer: Probabilistic Graphical Model explanations for Graph Neural Networks. arXiv [csLG] 2020; 12225–12235 Available from: https://proceedings.neurips.cc/paper/2020/hash/8fb134f258b1f7865a6ab2d935a897c9-Abstract.html

88 *Yuan H, Tang J, Hu X, Ji S.* XGNN: Towards Model-Level Explanations of Graph Neural Networks. In: Proceedings of the 26th ACM SIGKDD International Conference on Knowledge Discovery & Data Mining. New York, NY, USA: Association for Computing Machinery, 2020: 430–438

89 *Debnath AK, Lopez de Compadre RL, Debnath G, Shusterman AJ, Hansch C.* Structure-activity relationship of mutagenic aromatic and heteroaromatic nitro compounds. Correlation with molecular orbital energies and hydrophobicity. J Med Chem 1991; 34: 786–797 90 *Deng L.* The MNIST Database of Handwritten Digit Images for Machine Learning Research [Best of the Web]. IEEE Signal Process Mag 2012; 29: 141–142

91 *Agarwal C, Queen O, Lakkaraju H, Zitnik M.* Evaluating explainability for graph neural networks. Sci Data 2023; 10: 144

92 *Hu W, Cao K, Huang K, Huang EW, Subbian K, Leskovec J.* TuneUp: A Training Strategy for Improving Generalization of Graph Neural Networks. arXiv [statML] 2022; Available from: http://arxiv.org/abs/2210.14843

93 *Wang Q, Huang K, Chandak P, Zitnik M, Gehlenborg N.* Extending the Nested Model for User-Centric XAI: A Design Study on GNN-based Drug Repurposing. IEEE Trans Vis Comput Graph 2022; PP Available from: http://dx.doi.org/10.1109/TVCG.2022.3209435

94 *Yan Y, Hashemi M, Swersky K, Yang Y, Koutra D.* Two Sides of the Same Coin: Heterophily and Oversmoothing in Graph Convolutional Neural Networks. arXiv [csLG] 2021; Available from: http://arxiv.org/abs/2102.06462

95 *Hu Z, Dong Y, Wang K, Sun Y.* Heterogeneous Graph Transformer. In: Proceedings of The Web Conference 2020. New York, NY, USA: Association for Computing Machinery, 2020: 2704–2710

96 *Wang X, Ji H, Shi C, Wang B, Ye Y, Cui P, Yu PS.* Heterogeneous Graph Attention Network. In: The World Wide Web Conference. New York, NY, USA: Association for Computing Machinery, 2019: 2022–2032

97 *Frasca F, Rossi E, Eynard D, Chamberlain B, Bronstein M, Monti F.* SIGN: Scalable Inception Graph Neural Networks. arXiv [csLG] 2020; Available from: http://arxiv.org/abs/2004.11198

98 *Cui G, Zhou J, Yang C, Liu Z.* Adaptive Graph Encoder for Attributed Graph Embedding. arXiv [csLG] 2020; Available from: http://arxiv.org/abs/2007.01594

99 *Zeng H, Zhou H, Srivastava A, Kannan R, Prasanna V.* GraphSAINT: Graph Sampling Based Inductive Learning Method. arXiv [csLG] 2019; Available from: http://arxiv.org/abs/1907.04931

100 *Wang J, Ma A, Chang Y, Gong J, Jiang Y, Qi R, Wang C, Fu H, Ma Q, Xu D.* scGNN is a novel graph neural network framework for single-cell RNA-Seq analyses. Nat Commun 2021; 12: 1882

101 *Fu X, Zhang J, Meng Z, King I.* MAGNN: Metapath Aggregated Graph Neural Network for Heterogeneous Graph Embedding. arXiv [csSI] 2020; Available from: http://arxiv.org/abs/2002.01680

102 *Hu Z, Dong Y, Wang K, Sun Y.* Heterogeneous Graph Transformer. arXiv [csLG] 2020; Available from: http://arxiv.org/abs/2003.01332

103 *Zhu Y, Xu Y, Yu F, Liu Q, Wu S, Wang L.* Graph Contrastive Learning with Adaptive Augmentation. arXiv [csLG] 2020; Available from: http://arxiv.org/abs/2010.14945

104 *Shi C, Xu M, Zhu Z, Zhang W, Zhang M, Tang J.* GraphAF: a Flow-based Autoregressive Model for Molecular Graph Generation. arXiv [csLG] 2020; Available from: http://arxiv.org/abs/2001.09382

105 *Du Y, Guo X, Cao H, Ye Y, Zhao L.* Disentangled Spatiotemporal Graph Generative Models. AAAI 2022; 36: 6541–6549